\documentclass[11pt,a4paper]{article}
\usepackage[hyperref]{acl2017}
\usepackage{times}
\usepackage{latexsym}
\usepackage{graphicx}
\usepackage{amsmath}
\usepackage{amsfonts}
\usepackage{makecell}
\usepackage{tabularx}
\usepackage{url}

\graphicspath{ {images/} }

\aclfinalcopy 
\def\aclpaperid{***} 

\title{Massive Exploration of Neural Machine Translation Architectures}

\author{Denny Britz\thanks{Both authors contributed equally to this work.} \thanks{Work done as a member of the Google Brain Residency program (\texttt{g.co/brainresidency}).}, {Anna Goldie\footnotemark[1]}, {Minh-Thang Luong}, {Quoc Le} \\
\texttt{\{dennybritz,agoldie,thangluong,qvl\}@google.com} \\
Google Brain \\
}

\date{}

\begin{document}
\maketitle
\begin{abstract}
    Neural Machine Translation (NMT) has shown remarkable progress over the 
    past few years with production systems now being deployed 
    to end-users. One major drawback of current architectures
    is that they are expensive to train, typically requiring days to weeks of GPU
    time to converge. This makes exhaustive hyperparameter search,
    as is commonly done with other neural network architectures, prohibitively expensive. In this work, we present the first large-scale analysis of
    NMT architecture hyperparameters. We report empirical
    results and variance numbers for several hundred experimental runs, corresponding to over 250,000 GPU hours on the standard WMT English to German translation task. Our experiments lead to novel insights and practical advice for building and extending NMT architectures. As part of this contribution, we release an open-source NMT framework\footnote{https://github.com/google/seq2seq/} that enables researchers to easily experiment with novel techniques and reproduce state of the art results.
\end{abstract}

\section{Introduction}
\label{sec:introduction}

Neural Machine Translation (NMT) \cite{kal13,Sutskever:2014,Cho:2014} is an end-to-end approach to automated translation. NMT has shown impressive results \cite{Jean:2014, luong15acl,Sennrich:2016,Wu:2016} surpassing those of phrase-based systems while addressing shortcomings such as the need for hand-engineered features. The most popular approaches to NMT are based on an encoder-decoder architecture consisting of two recurrent neural networks (RNNs) and an attention mechanism that aligns target with source tokens \cite{Bahdanau:2014,Luong:2015}.

One shortcoming of current NMT architectures is the amount of compute required to train them. Training on real-world datasets of several million examples typically requires dozens of GPUs and convergence time is on the order of days to weeks \cite{Wu:2016}.  While sweeping across large hyperparameter spaces is common in Computer Vision \cite{Huang:2016b}, such exploration would be prohibitively expensive for NMT models, limiting researchers to well-established architectures and hyperparameter choices. Furthermore, there have been no large-scale studies of how architectural hyperparameters affect the performance of NMT systems. As a result, it remains unclear why these models perform as well as they do, as well as how we might improve them.

In this work, we present the first comprehensive analysis of architectural hyperparameters for Neural Machine Translation systems. Using a total of more than 250,000 GPU hours, we explore common variations of NMT architectures and provide insight into which architectural choices matter most. We report BLEU scores, perplexities, model sizes, and convergence time for all experiments, including variance numbers calculated across several runs of each experiment. In addition, we release to the public a new software framework that was used to run the experiments.

In summary, the main contributions of this work are as follows:

\begin{itemize}
\item We provide immediately applicable insights into the optimization of Neural Machine Translation models, as well as promising directions for future research. For example, we found that deep encoders are more difficult to optimize than decoders, that dense residual connections yield better performance than regular residual connections, that LSTMs outperform GRUs, and that a well-tuned beam search is crucial to obtaining state of the art results. By presenting practical advice for choosing baseline architectures, we help researchers avoid wasting time on unpromising model variations. 

\item We also establish the extent to which metrics such as BLEU are influenced by random initialization and slight hyperparameter variation, helping researchers to distinguish statistically significant results from random noise.

\item Finally, we release an open source package based on TensorFlow, specifically designed for implementing reproducible state of the art sequence-to-sequence models. All experiments were run using this framework and we hope to accelerate future research by releasing it to the public. We also release all configuration files and processing scripts needed to reproduce the experiments in this paper.

\end{itemize}

\section{Background and Preliminaries}
\label{sec:background}

\subsection{Neural Machine Translation}
\label{sec:nmt}

\begin{figure*}
\centering
\includegraphics[width=1.0\textwidth]{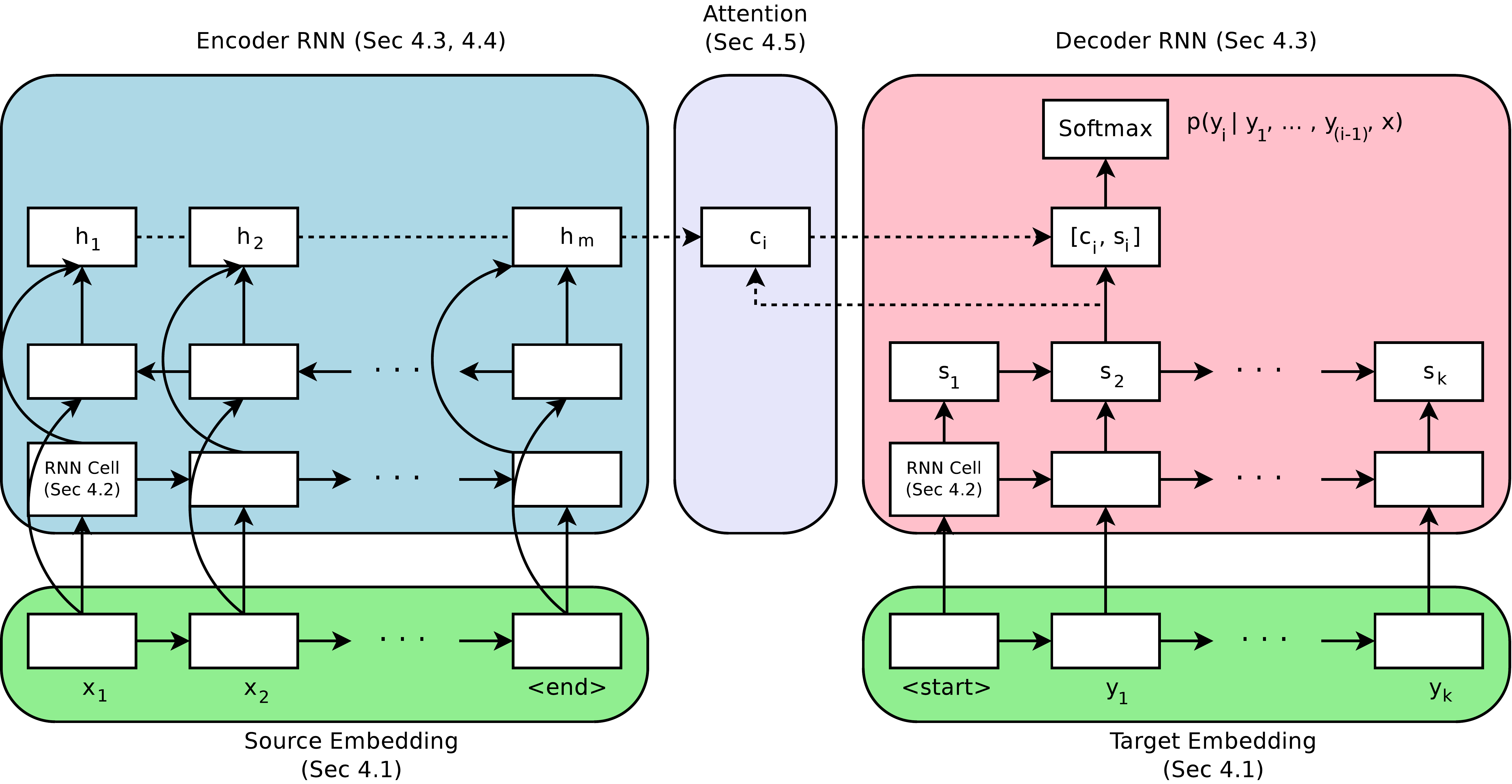}
\caption{Encoder-Decoder architecture with attention module. Section numbers reference experiments corresponding to the components.}
\label{fig:architecture}
\end{figure*}

Our models are based on an encoder-decoder architecture with attention mechanism \cite{Bahdanau:2014,Luong:2015}, as shown in figure \ref{fig:architecture}. An encoder function $f_{enc}$ takes as input a sequence of source tokens $\mathbf{x} = (x_1, ..., x_m)$ and produces a sequence of states $\mathbf{h} = (h_1, ..., h_m)$. In our base model, $f_{enc}$ is a bi-directional RNN and the state $h_i$ corresponds to the concatenation of the states produced by the backward and forward RNNs, $h_i = [\overrightarrow{h_i}; \overleftarrow{h_i}]$ .The decoder $f_{dec}$ is an RNN that predicts the probability of a target sequence $\mathbf{y} = (y_1, ..., y_k)$ based on $\mathbf{h}$. The probability of each target token $y_i \in {1, ... V}$ is predicted based on the recurrent state in the decoder RNN $s_i$, the previous words, $y_{<i}$, and a context vector $c_i$. The context vector $c_i$ is also called the attention vector and is calculated as a weighted average of the source states.

\begin{align}
c_i & = \sum_{j}{a_{ij} h_j} \\
a_{ij} & = \frac{\hat{a}_{ij}}{ \sum_j{\hat{a}_{ij}}} \\
\hat{a}_{ij} & = att(s_i, h_j)
\end{align}

Here, $att(s_i, h_j)$ is an attention function that calculates an unnormalized alignment score between the encoder state $h_j$ and the decoder state $s_i$. In our base model, we use a function of the form $att(s_i, h_j) = \langle W_hh_j,W_ss_i\rangle$, where the matrices $W$ are used to transform the source and target states into a representation of the same size.

The decoder outputs a distribution over a vocabulary of fixed-size $V$:

\begin{align*}
P(y_i \vert y_1, ..., y_{i-1}, \mathbf{x}) \\
= \text{softmax}(W[s_i; c_i] + b)
\end{align*}

The whole model is trained end-to-end by minimizing the negative log likelihood of the target words using stochastic gradient descent.

\section{Experimental Setup}
\label{sec:experimental_setup}

\subsection{Datasets and Preprocessing}
\label{sec:datasets}

We run all experiments on the WMT'15 English$\to$German task consisting of 4.5M sentence pairs, obtained by combining the Europarl v7, News Commentary v10, and Common Crawl corpora. We use newstest2013 as our validation set and newstest2014 and newstest2015 as our test sets. To test for generality, we also ran a small number of experiments on English$\to$French translation, and we found that the performance was highly correlated with that of English$\to$German but that it took much longer to train models on the larger English$\to$French dataset. Given that translation from the morphologically richer German is also considered a more challenging task, we felt justified in using the English$\to$German translation task for this hyperparameter sweep.

We tokenize and clean all datasets with the scripts in Moses\footnote{https://github.com/moses-smt/mosesdecoder/} and learn shared subword units using Byte Pair Encoding (BPE) \cite{Sennrich:2015} using 32,000 merge operations for a final vocabulary size of approximately 37k. We discovered that data preprocessing can have a large impact on final numbers, and since we wish to enable reproducibility, we release our data preprocessing scripts together with the NMT framework to the public. For more details on data preprocessing parameters, we refer the reader to the code release.

\subsection{Training Setup and Software}
\label{sec:training}

All of the following experiments are run using our own software framework based on TensorFlow \cite{Tensorflow:2016}. We purposely built this framework to enable reproducible state-of-the-art implementations of Neural Machine Translation architectures. As part of our contribution, we are releasing the framework and all configuration files needed to reproduce our results. Training is performed on Nvidia Tesla K40m and Tesla K80 GPUs, distributed over 8 parallel workers and 6 parameter servers per experiment. We use a batch size of 128 and decode using beam search with a beam width of 10 and the length normalization penalty of 0.6 described in \cite{Wu:2016}. BLEU scores are calculated on tokenized data using the \textit{multi-bleu.perl} script in Moses\footnote{https://github.com/moses-smt/mosesdecoder/blob/master/scripts/generic/multi-bleu.perl}. Each experiment is run for a maximum of 2.5M steps and replicated 4 times with different initializations. We save model checkpoints every 30 minutes and choose the best checkpoint based on the validation set BLEU score. We report mean and standard deviation as well as highest scores (as per cross validation) for each experiment.

\subsection{Baseline Model}
\label{sec:baseline}

Based on a review of previous literature, we chose a baseline model that we knew would perform reasonably well. Our goal was to keep the baseline model simple and standard, not to advance the start of the art. The model (described in \ref{sec:nmt}) consists of a 2-layer bidirectional encoder (1 layer in each direction), and a 2 layer decoder with a multiplicative \cite{Luong:2015} attention mechanism. We use 512-unit GRU \cite{Cho:2014} cells for both the encoder and decoder and apply Dropout of 0.2 at the input of each cell. We train using the Adam optimizer and a fixed learning rate of $0.0001$ without decay. The embedding dimensionality is set to 512. A more detailed description of all model hyperparameters can be found in the supplementary material.

In each of the following experiments, the hyperparameters of the baseline model are held constant, except for the one hyperparameter being studied. We hope that this allows us to isolate the effect of various hyperparameter changes. We recognize that this procedure does not account for interactions between hyperparameters, and we perform additional experiments when we believe such interactions are likely to occur (e.g. skip connections and number of layers).

\section{Experiments and Discussion}

For the sake of brevity, we only report mean BLEU, standard deviation, highest BLEU in parantheses, and model size in the following tables. Log perplexity, tokens/sec and convergence times can be found in the supplementary material tables.

\subsection{Embedding Dimensionality}
\label{sec:embedding}

With a large vocabulary, the embedding layer can account for a large fraction of the model parameters. Historically, researchers have used 620-dimensional \cite{Bahdanau:2014} or 1024-dimensional \cite{Luong:2015} embeddings. We expected larger embeddings to result in better BLEU scores, or at least lower perplexities, but we found that this wasn't always the case. While Table ~\ref{tab:embedding_dim} shows that \textit{2048-dimensional embeddings yielded the overall best result, they only did so by a small margin}. Even small 128-dimensional embeddings performed surprisingly well, while converging almost twice as quickly. We found that gradient updates to both small and large embeddings did not differ significantly and that the norm of gradient updates to the embedding matrix stayed approximately constant throughout training regardless of size. We also did not observe overfitting with large embeddings and training log perplexity was approximately equal across experiments, suggesting that the model does not make efficient use of the extra parameters and that there may be a need for better optimization techniques. Alternatively, it could be the case that models with large embeddings simply need much more than 2.5M steps to converge to the best solution.

\begin{table}[h]
\begin{center}
\begin{tabular}{|l|l|l|}
\hline \bf Dim & \bf newstest2013 & \bf Params \\ \hline
128 & $21.50 \pm 0.16$ (21.66) & 36.13M  \\
256 & $21.73 \pm 0.09$ (21.85)  & 46.20M  \\
512 & $21.78 \pm 0.05$ (21.83) & 66.32M  \\
1024 & $21.36 \pm 0.27$ (21.67) & 106.58M \\
2048 & $\mathbf{21.86} \pm 0.17$ (22.08) & 187.09M \\
\hline
\end{tabular}
\end{center}
\caption{\label{tab:embedding_dim} BLEU scores on newstest2013, varying the embedding dimensionality. }
\end{table}

\subsection{RNN Cell Variant}
\label{sec:cell}

Both LSTM \cite{LSTM:1997} and GRU \cite{Cho:2014} cells are commonly used in NMT architectures. While there exist studies \cite{Greff:2015} that explore cell variants on small sequence tasks of a few thousand examples, we are not aware of such studies in large-scale NMT settings.

A motivation for gated cells such as the GRU and LSTM is the vanishing gradient problem. Using vanilla RNN cells, deep networks cannot efficiently propagate information and gradients through multiple layers and time steps. However, with an attention-based model, we believe that the decoder should be able to make decisions almost exclusively based on the current input and the attention context and we hypothesize that the gating mechanism in the decoder is not strictly necessary. This hypothesis is supported by the fact that we always initialize the decoder state to zero instead of passing the encoder state, meaning that the decoder state does not contain information about the encoded source. We test our hypothesis by using a vanilla RNN cell in the decoder only (Vanilla-Dec below). For the LSTM and GRU variants we replace cells in both the encoder and decoder. We use LSTM cells without peephole connections and initialize the forget bias of both LSTM and GRU cells to 1.

\begin{table}[h]
\begin{center}
\begin{tabular}{|l|l|l|}
\hline \bf Cell & \bf newstest2013 & \bf Params \\ \hline
LSTM & $\mathbf{22.22} \pm 0.08$ (22.33)  & 68.95M \\
GRU  & $21.78 \pm 0.05$ (21.83) & 66.32M  \\
Vanilla-Dec & $15.38 \pm 0.28$ (15.73)  & 63.18M \\
\hline
\end{tabular}
\end{center}
\caption{\label{font-table} BLEU scores on newstest2013, varying the type of encoder and decoder cell. }
\end{table}

In our experiments, \textit{LSTM cells consistently outperformed GRU cells}. Since the computational bottleneck in our architecture is the softmax operation we did not observe large difference in training speed between LSTM and GRU cells. Somewhat to our surprise, we found that the vanilla decoder is unable to learn nearly as well as the gated variant. This suggests that the decoder indeed passes information in its own state throughout multiple time steps instead of relying solely on the attention mechanism and current input (which includes the previous attention context). It could also be the case that the gating mechanism is necessary to mask out irrelevant parts of the inputs.

\subsection{Encoder and Decoder Depth}
\label{sec:depth}

We generally expect deeper networks to converge to better solutions than shallower ones \cite{He:2015}. While some work \cite{luong15acl, Zhou:2016, luong16acl, Wu:2016} has achieved state of the art results using deep networks, others \cite{Jean:2014, Chung:2016, Sennrich:2015} have achieved similar results with far shallower ones. Hence, it is unclear how important depth is, and whether shallow networks are capable of producing results competitive with those of deep networks. Here, we explore the effect of both encoder and decoder depth up to 8 layers. For the bidirectional encoder, we separately stack the RNNs in both directions. For example, the Enc-8 model corresponds to one forward and one backward 4-layer RNN. For deeper networks, we also experiment with two variants of residual connections \cite{He:2015} to encourage gradient flow. In the standard variant, shown in equation \eqref{eq:residual:standard}, we insert residual connections between consecutive layers. If $h_t^{(l)}(x_t^{(l)}, h_{t-1}^{(l)})$ is the RNN output of layer $l$ at time step $t$, then:

\begin{align}
\label{eq:residual:standard}
x_t^{(l+1)} &= h_t^{(l)}(x_t^{(l)}, h_{t-1}^{(l)}) + x_t^{(l)}
\end{align}

where $x_t^{(0)}$ are the embedded input tokens. We also explore a dense ("ResD" below) variant of residual connections similar to those used by \cite{Huang:2016} in Image Recognition. In this variant, we add skip connections from each layer to all other layers:

\begin{align}
\label{eq:residual:dense}
x_t^{(l+1)} &= h_t^{(l)}(x_t^{(l)}, h_{t-1}^{(l)}) + \sum_{j=0}^l{x_t^{(j)}}
\end{align}

Our implementation differs from \cite{Huang:2016} in that we use an addition instead of a concatenation operation in order to keep the state size constant.

\begin{figure}
\centering
\includegraphics[width=7.7cm]{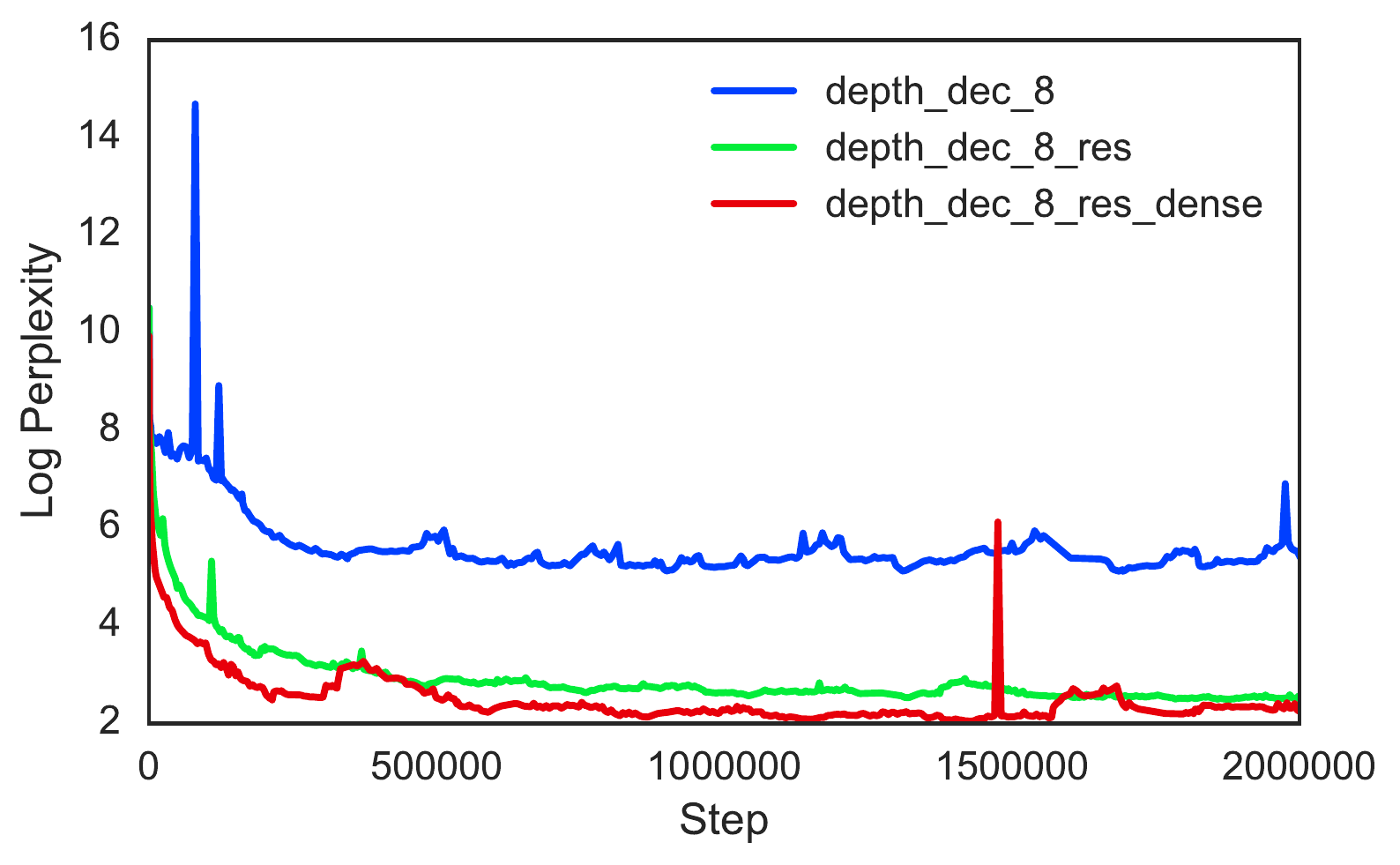}
\caption{Training plots for deep decoder with and without residual connections, showing log perplexity on the eval set.}
\label{fig:deep_residual}
\end{figure}

\begin{table}[h]
\begin{center}
\begin{tabular}{|l|l|l|}
\hline \bf Depth & \bf newstest2013 & \bf Params \\ \hline
Enc-2 & $21.78 \pm 0.05$ (21.83) & 66.32M  \\
Enc-4 & $\mathbf{21.85} \pm 0.32$ (22.23) & 69.47M \\
Enc-8 & $21.32 \pm 0.14$ (21.51) & 75.77M \\
Enc-8-Res & $19.23 \pm 1.96$ (21.97) & 75.77M \\
Enc-8-ResD & $17.30 \pm 2.64$ (21.03) & 75.77M \\
\hline
Dec-1 & $21.76 \pm 0.12$ (21.93) & 64.75M \\
Dec-2 & $21.78 \pm 0.05$ (21.83) & 66.32M  \\
Dec-4 & $\mathbf{22.37} \pm 0.10$ (22.51) & 69.47M \\
Dec-4-Res & $17.48 \pm 0.25$ (17.82) & 68.69M \\
Dec-4-ResD & $21.10 \pm 0.24$ (21.43) & 68.69M \\
Dec-8 & $01.42 \pm 0.23$ (1.66) & 75.77M \\
Dec-8-Res & $16.99 \pm 0.42$ (17.47) & 75.77M \\
Dec-8-ResD & $20.97 \pm 0.34$ (21.42) & 75.77M \\
\hline
\end{tabular}
\end{center}
\caption{\label{table:depth} BLEU scores on newstest2013, varying the encoder and decoder depth and type of residual connections. }
\end{table}

Table \ref{table:depth} shows results of varying encoder and decoder depth with and without residual connection. We found no clear evidence that encoder depth beyond two layers is necessary, but found deeper models with residual connections to be significantly more likely to diverge during training. The best deep residual models achieved good results, but only one of four runs converged, as suggested by the large standard deviation. 

On the decoder side, deeper models outperformed shallower ones by a small margin, and we found that without residual connections, it was impossible for us to train decoders with 8 or more layers. Across the deep decoder experiments, dense residual connections consistently outperformed regular residual connections and converged much faster in terms of step count, as shown in figure \ref{fig:deep_residual}. We expected deep models to perform better \cite{Zhou:2016, Szegedy:2014} across the board, and we believe that our experiments demonstrate the need for more robust techniques for optimizing deep sequential models. For example, we may need a better-tuned SGD optimizer or some form of batch normalization, in order to robustly train deep networks with residual connections.

\subsection{Unidirectional vs. Bidirectional Encoder}
\label{sec:encoder}

In the literature, we see bidirectional encoders \cite{Bahdanau:2014}, unidirectional encoders \cite{Luong:2015}, and a mix of both \cite{Wu:2016} being used. Bidirectional encoders are able to create representations that take into account both past and future inputs, while unidirectional encoders can only take past inputs into account. The benefit of unidirectional encoders is that their computation can be easily parallelized on GPUs, allowing them to run faster than their bidirectional counterparts. We are not aware of any studies that explore the necessity of bidirectionality. In this set of experiments, we explore unidirectional encoders of varying depth with and without reversed source inputs, as this is a commonly used trick that allows the encoder to create richer representations for earlier words. Given that errors on the decoder side can easily cascade, the correctness of early words has disproportionate impact.

\begin{table}[h]
\begin{center}
\begin{tabular}{|l|l|l|}
\hline \bf Cell & \bf newstest2013  & \bf Params \\ \hline
Bidi-2 & $\mathbf{21.78} \pm 0.05$ (21.83) & 66.32M  \\
Uni-1 & $20.54 \pm 0.16$ (20.73) & 63.44M \\
Uni-1R & $21.16 \pm 0.35$ (21.64) & 63.44M \\
Uni-2 & $20.98 \pm 0.10$ (21.07) & 65.01M \\
Uni-2R & $21.76 \pm 0.21$ (21.93) & 65.01M \\
Uni-4 & $21.47 \pm 0.22$ (21.70) & 68.16M \\
Uni-4R & $21.32 \pm 0.42$ (21.89) & 68.16M \\
\hline
\end{tabular}
\end{center}
\caption{\label{tab:encoder} BLEU scores on newstest2013, varying the type of encoder. The "R" suffix indicates a reversed source sequence. }
\end{table}

Table \ref{tab:encoder} shows that bidirectional encoders generally outperform unidirectional encoders, but not by a large margin. The encoders with reversed source consistently outperform their non-reversed counterparts, but do not beat shallower bidirectional encoders.

\subsection{Attention Mechanism}
\label{sec:attention}

The two most commonly used attention mechanisms are the additive \cite{Bahdanau:2014} variant, equation \eqref{eq:attention:bahdanau} below, and the computationally less expensive multiplicative variant \cite{Luong:2015}, equation \eqref{eq:attention:dot} below. Given an attention key $h_j$ (an encoder state) and attention query $s_i$ (a decoder state), the attention score for each pair is calculated as follows:

\begin{align}
\label{eq:attention:bahdanau}
\text{score}(h_j, s_i) &= \langle v, tanh(W_1 h_j + W_2 s_i)\rangle \\
\label{eq:attention:dot}
\text{score}(h_j, s_i) &=  \langle W_1 h_j, W_2 s_i \rangle
\end{align}

We call the dimensionality of $W_1 h_j$ and $W_2 s_i$ the "attention dimensionality" and vary it from 128 to 1024 by changing the layer size. We also experiment with using no attention mechanism by initializing the decoder state with the last encoder state (None-State), or concatenating the last decoder state to each decoder input (None-Input). The results are shown in Table \ref{table:attention}.

\begin{table}[h]
\begin{center}
\begin{tabular}{|l|l|l|}
\hline \bf Attention & \bf newstest2013  & \bf Params \\ \hline
Mul-128 & $22.03 \pm 0.08$ (22.14) & 65.73M \\
Mul-256 & $22.33 \pm 0.28$ (22.64) & 65.93M \\
Mul-512 & $21.78 \pm 0.05$ (21.83) & 66.32M  \\
Mul-1024 & $18.22 \pm 0.03$ (18.26) & 67.11M \\
Add-128 & $22.23 \pm 0.11$ (22.38) & 65.73M \\
Add-256 & $22.33 \pm 0.04$ (22.39) & 65.93M \\
Add-512 & $\mathbf{22.47} \pm 0.27$ (22.79) & 66.33M \\
Add-1028 & $22.10 \pm 0.18$ (22.36) & 67.11M \\
None-State & $9.98 \pm 0.28$ (10.25) & 64.23M \\
None-Input & $11.57 \pm 0.30$ (11.85) & 64.49M \\
\hline
\end{tabular}
\end{center}
\caption{\label{table:attention} BLEU scores on newstest2013, varying the type of attention mechanism. }
\end{table}

We found that the parameterized additive attention mechanism slightly but consistently outperformed the multiplicative one, with the attention dimensionality having little effect.

While we did expect the attention-based models to significantly outperform those without an attention mechanism, we were surprised by just how poorly the "Non-Input" models fared, given that they had access to encoder information at each time step. Furthermore, we found that the attention-based models exhibited significantly larger gradient updates to decoder states throughout training. This suggests that the attention mechanism acts more like a "weighted skip connection" that optimizes gradient flow than like a "memory" that allows the encoder to access source states, as is commonly stated in the literature. We believe that further research in this direction is necessary to shed light on the role of the attention mechanism and whether it may be purely a vehicle for easier optimization.

\subsection{Beam Search Strategies}
\label{sec:beam}

Beam Search is a commonly used technique to find target sequences that maximize some scoring function $s(\mathbf{y}, \mathbf{x})$ through tree search. In the simplest case, the score to be maximized is the log probability of the target sequence given the source. Recently, extensions such as coverage penalties \cite{Tu:2016} and length normalizations \cite{Wu:2016} have been shown to improve decoding results. It has also been observed \cite{Tu:2016b} that very large beam sizes, even with length penalty, perform worse than smaller ones. Thus, choosing the correct beam width can be crucial to achieving the best results.

\begin{table}[h]
\begin{center}
\begin{tabular}{|l|l|l|}
\hline \bf Beam & \bf newstest2013  & \bf Params \\ \hline
B1 & $20.66 \pm 0.31$ (21.08) & 66.32M  \\
B3 & $21.55 \pm 0.26$ (21.94) & 66.32M  \\
B5 & $21.60 \pm 0.28$ (22.03) & 66.32M  \\
B10 & $21.57 \pm 0.26$ (21.91) & 66.32M  \\
B25 & $21.47 \pm 0.30$ (21.77) & 66.32M  \\
B100 & $21.10 \pm 0.31$ (21.39) & 66.32M  \\
B10-LP-0.5 & $21.71 \pm 0.25$ (22.04) & 66.32M  \\
B10-LP-1.0 & $\mathbf{21.80} \pm 0.25$ (22.16) & 66.32M \\
\hline
\end{tabular}
\end{center}
\caption{\label{table:beam} BLEU scores on newstest2013, varying the beam width and adding length penalties (LP). }
\end{table}

Table \ref{table:beam} shows the effect of varying beam widths and adding length normalization penalties. A beam width of 1 corresponds to greedy search. We found that a well-tuned beam search is crucial to achieving good results, and that it leads to consistent gains of more than one BLEU point. Similar to \cite{Tu:2016b} we found that very large beams yield worse results and that there is a "sweet spot" of optimal beam width. We believe that further research into the robustness of hyperparameters in beam search is crucial to progress in NMT. We also experimented with a coverage penalty, but found no additional gain over a sufficiently large length penalty.

\subsection{Final System Comparison}

Finally, we compare our best performing model across all experiments (base model with 512-dimensional additive attention), as chosen on the newstest2013 validation set, to historical results found in the literature in Table \ref{table:results}. While not the focus on this work, we were able to achieve further improvements by combining all of our insights into a single model described in Table \ref{table:combined}.

\begin{table}[h]
\begin{center}
\begin{tabular}{|l|l|l|}
\hline \bf Hyperparameter & \bf Value \bf \\ \hline
embedding dim & 512 \\
rnn cell variant & LSTMCell \\
encoder depth & 4 \\
decoder depth & 4 \\
attention dim & 512 \\
attention type & Bahdanau \\
encoder & bidirectional \\
beam size & 10 \\
length penalty & 1.0 \\
\hline
\end{tabular}
\end{center}
\caption{\label{table:combined} Hyperparameter settings for our final combined model, consisting of all of the individually optimized values.}
\end{table}

Although we do not offer architectural innovations, we do show that through careful hyperparameter tuning and good initialization, it is possible to achieve state of the art performance on standard WMT benchmarks. Our model is outperformed only by \cite{Wu:2016}, a model which is significantly more complex and lacks a public implementation.

\begin{table}[h!]
\begin{center}
\begin{tabular}{|l|l|l|}
\hline \bf Model & \bf newstest14 & \bf newstest15 \\ \hline
Ours (experimental) & 22.03 & 24.75 \\
Ours (combined) & 22.19 & 25.23 \\
\hline
OpenNMT & 19.34 & - \\
Luong & 20.9 & - \\
BPE-Char & 21.5 & 23.9 \\
BPE & - & 20.5 \\
RNNSearch-LV & 19.4 & - \\
RNNSearch & - & 16.5 \\
\hline
Deep-Att\textsuperscript{*} & 20.6 & -  \\
GNMT\textsuperscript{*} & 24.61 & - \\
Deep-Conv\textsuperscript{*} &-  & 24.3  \\
\hline
\end{tabular}
\end{center}
\caption{\label{table:results} Comparison to RNNSearch \cite{Jean:2014}, RNNSearch-LV \cite{Jean:2014}, BPE \cite{Sennrich:2015}, BPE-Char \cite{Chung:2016}, Deep-Att \cite{Zhou:2016}, Luong \cite{Luong:2015}, Deep-Conv \cite{Gehring:2016}, GNMT \cite{Wu:2016}, and OpenNMT \cite{Klein:2017}. Systems with an \textsuperscript{*} do not have a public implementation. }
\end{table}

\section{Open Source Release}

We demonstrated empirically how small changes to hyperparameter values and different initialization can affect results, and how seemingly trivial factors such as a well-tuned beam search are crucial. To move towards reproducible research, we believe it is important that researchers start building upon common frameworks and data processing pipelines. With this goal in mind, we specifically built a modular software framework that allows researchers to explore novel architectures with minimal code changes, and define experimental parameters in a reproducible manner. While our initial experiments are in Machine Translation, our framework can easily be adapted to problems in Summarization, Conversational Modeling or Image-To-Text. Systems such as OpenNMT \cite{Klein:2017} share similar goals, but do not yet achieve state of the art results (see Table \ref{table:results}) and lack what we believe to be crucial features, such as distributed training support. We hope that by open sourcing our experimental toolkit, we enable the field to make more rapid progress in the future.

All of our code is freely available at https://github.com/google/seq2seq/.

\section{Conclusion}

We conducted what we believe to be the first large-scale analysis of architecture variations for Neural Machine Translation, teasing apart the key factors to achieving state of the art results. We demonstrated a number of surprising insights, including the fact that beam search tuning is just as crucial as most architectural variations, and that with current optimization techniques deep models do not always outperform shallow ones. Here, we summarize our practical findings:

\begin{itemize}
\item Large embeddings with 2048 dimensions achieved the best results, but only by a small margin. Even small embeddings with 128 dimensions seem to have sufficient capacity to capture most of the necessary semantic information.
\item LSTM Cells consistently outperformed GRU Cells.
\item Bidirectional encoders with 2 to 4 layers performed best. Deeper encoders were significantly more unstable to train, but show potential if they can be optimized well.
\item Deep 4-layer decoders slightly outperformed shallower decoders. Residual connections were necessary to train decoders with 8 layers and dense residual connections offer additional robustness.
\item Parameterized additive attention yielded the overall best results.
\item A well-tuned beam search with length penalty is crucial. Beam widths of 5 to 10 together with a length penalty of 1.0 seemed to work well.
\end{itemize}

We highlighted several important research questions, including the efficient use of embedding parameters (\ref{sec:embedding}), the role of attention mechanisms as weighted skip connections (\ref{sec:attention}) as opposed to memory units, the need for better optimization methods for deep recurrent networks (\ref{sec:depth}), and the need for a better beam search (\ref{sec:beam}) robust to hyperparameter variations.

In addition, we release to the public an open source NMT framework specifically built to explore architectural innovations and generate reproducible experiments, along with configuration files for all our experiments.

\subsubsection*{Acknowledgments}

We would like to thank Eugene Brevdo for adapting the TensorFlow RNN APIs in a way that allowed us to write our framework much more cleanly. We are also grateful to Andrew Dai and Samy Bengio for their helpful feedback.

\bibliography{acl}
\bibliographystyle{acl_natbib}

\end{document}